\documentclass{article}
\usepackage{spconf,amsmath,graphicx}
\usepackage[acronym]{glossaries}
\usepackage{subfigure}
\usepackage{multirow}
\usepackage{amssymb}
\usepackage{bm}
\usepackage{color}
\newacronym{g-pcc}{G-PCC}{geometry-based point cloud compression}
\newacronym{v-pcc}{V-PCC}{video-based point cloud compression}
\newacronym{sgwpcqa}{SGW-PCQA}{spectral graph wavelet-based point cloud quality assessment}

\newacronym{pcqagsv}{PCQA-GSV}{point cloud quality assessment with graph signal variations}
\newacronym{dslg}{DSLG}{diagonal scan-line graph}
\newacronym{svr}{SVR}{support vector regression}
\newacronym{frsvr}{FRSVR}{full-reference quality assessment using support vector regression}

\newacronym{sgw}{SGW}{spectral graph wavelet}
\newacronym{mos}{MOS}{mean opinion score}
\newacronym{gtv}{GTV}{graph total variation}

\newacronym{pcqm}{PCQM}{point cloud quality metric}
\newacronym{basics}{BASICS}{broad quality assessment of static point clouds in compression scenario}
\newacronym{wpc}{WPC}{Waterloo point cloud}

\newacronym{sgwt}{SGWT}{spectral graph wavelet transform}

\newacronym{ngsv}{NGSV}{normalized graph signal variations}
\newacronym{knn}{KNN}{K nearest neighbors}
\newacronym{glr}{GLR}{graph Laplacian regularization}
\newacronym{cd-sgw}{CD-SGW}{spectral graph wavelet-based color denoising}
\newacronym{3dpbs}{3DPBS}{3-dimensional patch-based similarity}
\newacronym{lbvh}{LBVH}{linear bounding volume hierarchy}
\newacronym{fgbd}{FGBD}{fast graph-based denoising}
\newacronym{slg}{SLG}{scan-line graph}
\newacronym{ne-gbp}{NE-GBP}{noise estimation using graph-based patches}
\newacronym{fslr}{FSLR}{filter selection with limited region}
\newacronym{pca}{PCA}{principal component analysis}
\newacronym{mvub}{MVUB}{Microsoft voxelized upper bodies}
\newacronym{psnr}{PSNR}{Peak signal-to-noise ratio}
\newacronym{bf-knn}{BF-KNN}{Brute-force KNN}
\newacronym{megw}{MEGW}{median estimator with graph wavelets}
\newacronym{qa}{QA}{quality assessment}
\newacronym{pcqa}{PCQA}{point cloud quality assessment}
\newacronym{nr-pcqa}{NR-PCQA}{no-reference point cloud quality assessment}
\newacronym{nr}{NR}{no-reference}
\newacronym{fr-pcqa}{FR-PCQA}{full-reference point cloud quality assessment}

\usepackage{hyperref}

\title{FULL-REFERENCE POINT CLOUD QUALITY ASSESSMENT \\USING SPECTRAL GRAPH WAVELETS}
\name{
{
Ryosuke Watanabe $^{\dagger,\ddagger}$, 
Keisuke Nonaka $^{\ddagger}$,
Eduardo Pavez $^{\dagger}$,
Tatsuya Kobayashi $^{\ddagger}$, 
Antonio Ortega $^{\dagger}$
}
}
\address{
$^{\dagger}$ University of Southern California,
$^{\ddagger}$ KDDI Research, Inc.  
\thanks{
Copyright~\copyright~2024 IEEE. Personal use of this material is permitted. Permission from IEEE must be obtained for all other uses, in any current or future media, including reprinting/republishing this material for advertising or promotional purposes, creating new collective works, for resale or redistribution to servers or lists, or reuse of any copyrighted component of this work in other work.}
}

\begin{document}
%
\maketitle
\begin{abstract}
Point clouds in 3D applications frequently experience quality degradation during processing, e.g., scanning and compression.
Reliable point cloud quality assessment (PCQA) is important for developing compression algorithms with good bitrate-quality trade-offs and techniques for quality improvement (e.g., denoising).
This paper introduces a full-reference (FR) PCQA method utilizing spectral graph wavelets (SGWs).
First, we propose novel SGW-based PCQA metrics that compare SGW coefficients of coordinate and color signals between reference and distorted point clouds. 
Second, we achieve accurate PCQA by integrating several conventional FR metrics and our SGW-based metrics using support vector regression.
To our knowledge, this is the first study to introduce SGWs for PCQA. 
Experimental results demonstrate the proposed PCQA metric is more accurately correlated with subjective quality scores compared to conventional PCQA metrics.
\end{abstract}
%
%

%
%
\section{Introduction}
\label{sec:introduction}

Point clouds are a general 3D format representing realistic 3D objects in diverse 3D applications such as telepresence, monitoring, and holographic display.
However, the perceptual quality of point clouds often deteriorates during scanning, compression, and transmission.
Point cloud compression methods have been proposed to achieve a good trade-off between quality and bitrate.
For instance, the MPEG committee has standardized two approaches: \gls{g-pcc} and \gls{v-pcc} \cite{MPEGPCC}. 
Moreover, many methods aiming at improving point cloud quality have been proposed to mitigate the impact of distortions (e.g., point cloud denoising \cite{PCDenoising}, upsampling \cite{PCUpsampling}, and inpainting \cite{PCinpainting}).
Accurate \gls{pcqa} methods are essential to evaluate compression and quality enhancement methods.

Methods for \gls{pcqa} can be classified into \gls{fr-pcqa} and \gls{nr-pcqa} approaches. 
Unlike \gls{nr-pcqa} metrics \cite{PQANet,NRPCQA,DeepNR}, which cannot access a reference (noise-free) point cloud, \gls{fr-pcqa} methods show reliable and stable assessment results when a good reference is available.
For example, some of the \gls{fr-pcqa} metrics \cite{MPEGEval,C2PError} have been utilized as criteria for evaluating point cloud compression methods in the MPEG standardization \cite{MPEGPCC}.
While these methods \cite{MPEGEval,C2PError}  primarily focus on point-wise errors, alternative approaches that consider (i) more complex features (e.g., structural similarity~\cite{PointSSIM,PCQM,MSPointSSIM} or (ii) graph similarity~\cite{GraphSIM,MSGraphSIM}) have been proposed to improve the correlation with subjective assessment scores such as \gls{mos}.

In recent years, learning-based \gls{fr-pcqa} methods \cite{DeepFR,PointPCA+,ICIPChallenge} have been shown to obtain better correlation with subjective assessment scores.
Our previous study, \gls{frsvr} \cite{ICIPChallenge}, achieves high correlation with subjective assessment scores by integrating five different \gls{fr-pcqa} metrics using \gls{svr} and achieved first place in the FR broad-range quality estimation track in the ICIP 2023 point cloud visual quality assessment grand challenge (ICIP 2023 PCVQA grand challenge)~\cite{PCVQA2023}.
Since it was designed for a grand challenge focused on \gls{pcqa} for compression distortion only, \gls{frsvr} achieves less correlation with subjective assessment scores for other types of noise (e.g., Gaussian noise, down-sampling). 
In particular, the five types of scores adopted by \gls{frsvr} do not consider multi-scale features, which, 
according to previous studies \cite{MSPointSSIM,MSGraphSIM}, 
are effective in assessing various types of distortion.
Specifically, \gls{frsvr} utilizes two point-wise error metrics to evaluate geometric distortion but does not carry out region-to-region comparisons. 
Similarly, although region-to-region comparison is performed for color distortion, this is a single-scale feature calculated from a specified number of neighbors.

As an alternative, in this paper, we propose a more accurate \gls{fr-pcqa} method using a \gls{sgw}, which can provide multi-scale features as graph signal frequencies.
In addition, the proposed method is faster than conventional multi-scale \gls{fr-pcqa} methods \cite{MSPointSSIM, MSGraphSIM} because a fast calculation technique has been proposed for \gls{sgwt}s such as polynomial approximation \cite{SGWT}. 

Our contributions are as follows:
\begin{enumerate}
\item we propose a novel \gls{sgwpcqa} metric that compares \gls{sgw}s calculated from the coordinate and color
signals of both reference and distorted point clouds.
The previous studies on point cloud denoising based on \gls{sgw}s \cite{MSGW,GAC,SCNNK} demonstrate that noise in geometric and color information can be eliminated by suppressing the high-frequency components of \gls{sgw}s. 
Thus, comparing \gls{sgw}s calculated from reference and distorted point clouds is useful to evaluate the amount of noise in the distorted point cloud.
As far as we know, this is the first study to introduce \gls{sgw}s for \gls{pcqa}. 
\item To improve the correlation coefficients with subjective assessment scores, we use \gls{svr} to 
integrate the \gls{sgwpcqa} metrics with two out of the five metrics introduced in our previous \gls{svr}-based method, \gls{frsvr}~\cite{ICIPChallenge}.
This integrated version, \gls{sgwpcqa}+, leads to better correlation coefficients 
for both compression distortion and other types of distortions, such as Gaussian noise and downsampling. 
In addition, \gls{sgwpcqa}+ outperformed the results of our previous work (\gls{frsvr}~\cite{ICIPChallenge}) recorded in the ICIP 2023 PCVQA grand challenge in terms of the correlation coefficients with \gls{mos} and processing time.
\end{enumerate}

%
%
%
%
%
\section{Related work}
\label{sec:related-work}

\begin{figure}[t]
\centering 
\includegraphics[width=0.8\linewidth]{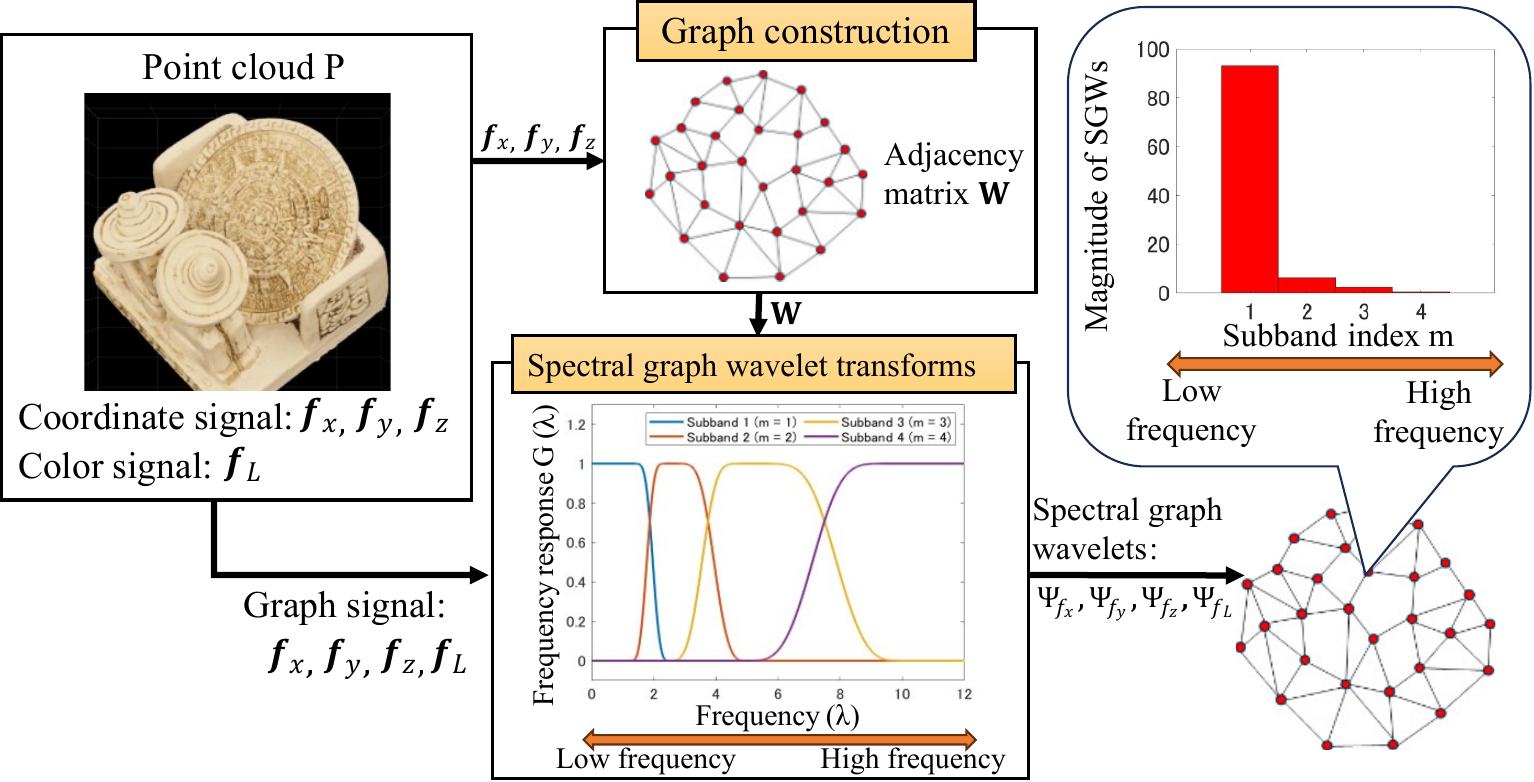}
   \caption{Calculation of spectral graph wavelet transforms (SGWTs) for a point cloud.}
\label{fig:sgwt}
\end{figure}
\begin{figure*}[t]
\centering 
\includegraphics[width=0.8\linewidth]{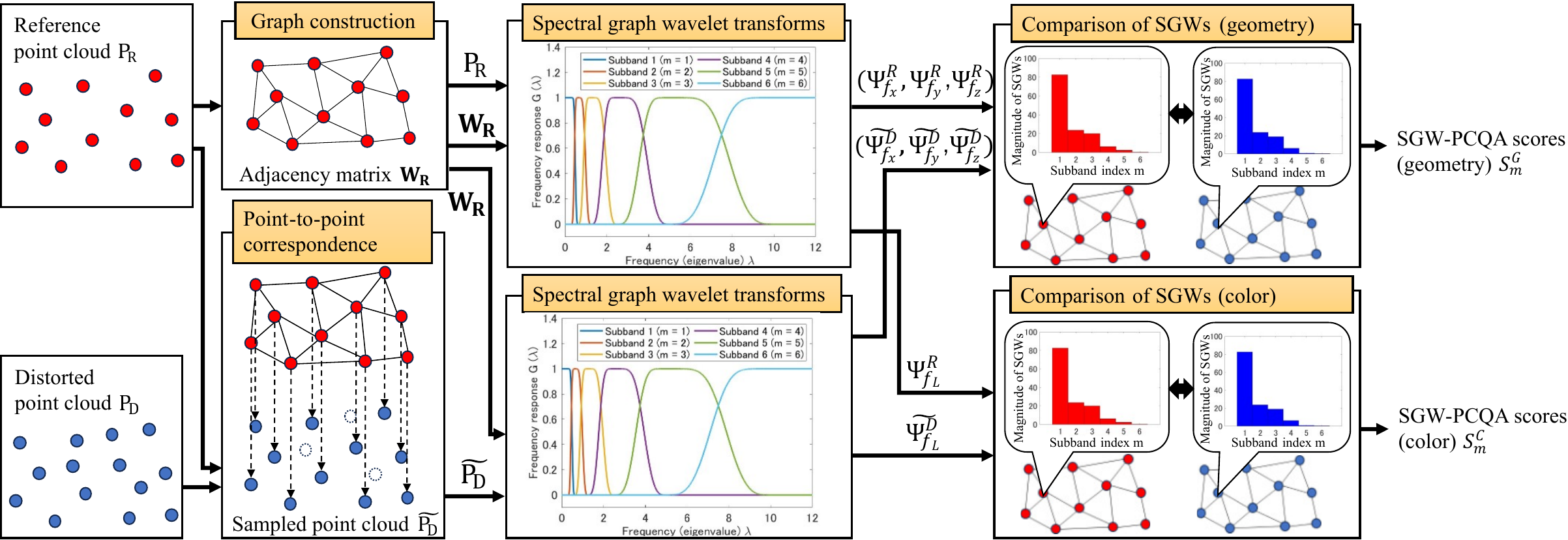}
   \caption{Overall calculation flow of the proposed \gls{pcqa} metrics using SGWs.}
\label{fig:flow}
\end{figure*}

Since this paper focuses on \gls{fr-pcqa}, we introduce conventional \gls{fr-pcqa} methods in this section.
\subsection{3D-to-2D projection-based methods}
\label{sub-sec:projection-based-method}
%
Conventional image quality assessment techniques can be used to assess the quality of 3D point clouds by projecting them into multiple 2D planes.
In \cite{Projection}, multiple projected 2D images are assessed by popular 2D image quality assessment metrics (e.g., VIFP~\cite{VIFP}, SSIM~\cite{SSIM}).
An attention-guided \gls{pcqa} inspired by the structural similarity measure of \cite{IW-SSIM} has also been proposed~\cite{IW-SSIMp}.
However, these methods suffer from a low correlation with subjective assessment scores since there is a loss of 3D information during projection.
\subsection{3D model-based methods}
\label{sub-sec:3d-based-method}
Geometric distortions, e.g., point-to-point error ~\cite{MPEGEval}, point-to-plane error~\cite{C2PError} and plane-to-plane error~\cite{AngularSimilarity}, have also been introduced, but they cannot evaluate color distortion, so their applicability for \gls{pcqa} is limited.
%

To evaluate color distortion, the Peak Signal-to-Noise Ratio (PSNR)~\cite{MPEGEval} and PointSSIM~\cite{PointSSIM} have been proposed. 
However, their correlation with subjective assessment scores is still limited since these methods compare only one feature (e.g., color \cite{MPEGEval} or local color variance \cite{PointSSIM}).
To improve the reliability of \gls{pcqa} metrics, \gls{pcqm}~\cite{PCQM} introduces a weighted linear combination of geometric and color distortions.
Graph-based \gls{fr-pcqa} distortions methods\cite{GraphSIM} and multi-scale methods have also been proposed to further improve the accuracy~\cite{MSPointSSIM,MSGraphSIM}.
However, the improved accuracy of \cite{PCQM,MSPointSSIM,GraphSIM,MSGraphSIM} 
comes with a large increase in computational complexity. 
\subsection{Learning-based methods}
Recently, learning-based techniques have been introduced to obtain better correlation with subjective assessment scores.
Unlike previously mentioned approaches, these methods require training a model using subjective assessment scores such as \gls{mos}. 
Examples of these approaches include~\cite{DeepFR}, an end-to-end deep-learning framework for accurate \gls{fr-pcqa} and \cite{PointPCA+}, 
 a learning-based method utilizing various PCA-based features. 
While \cite{DeepFR, PointPCA+} demonstrate a high correlation with subjective assessment scores and combine geometry and color information, they suffer from large processing time for prediction.
By using the \gls{svr}-based method \cite{ICIPChallenge} to combine simple \gls{fr-pcqa} metrics by \gls{svr}, we achieve high correlation with subjective assessment scores for compression distortion with relatively low complexity.
However, this approach is less reliable for other types of noises (e.g., Gaussian noise, down-sampling), as discussed in \autoref{sec:introduction}.

\begin{figure*}[t]
\centering 
   \subfigure[]{
   \includegraphics[width=.80\columnwidth]{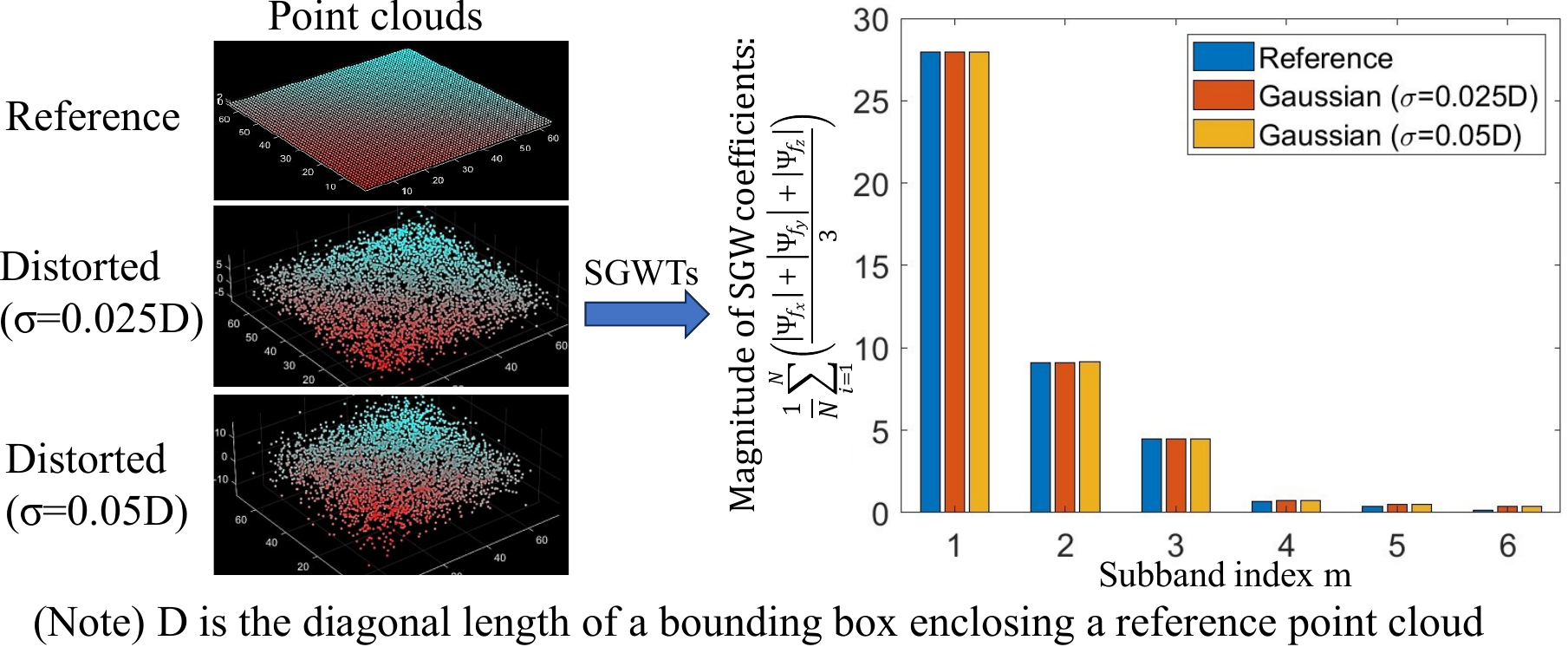}
   \label{fig:graphGaussianGeom}
  }
  \subfigure[]{
   \includegraphics[width=.80\columnwidth]{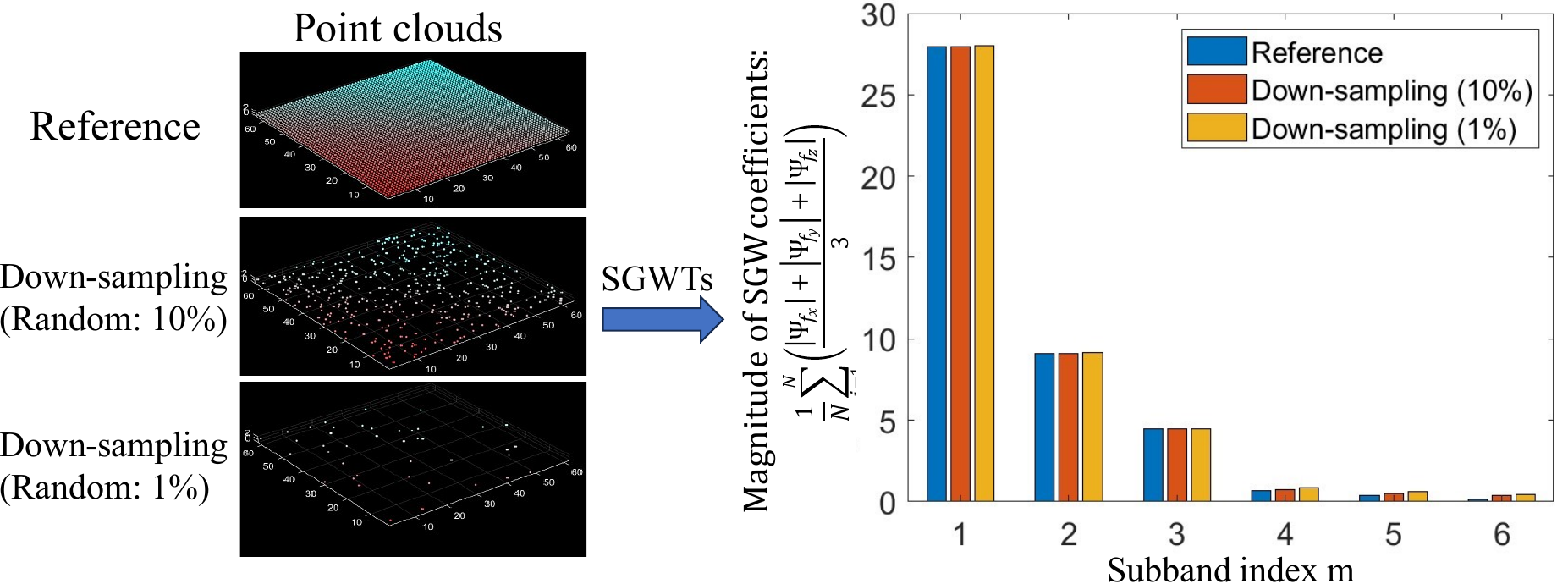}
   \label{fig:graphDownsampleGeom}
  }
    \subfigure[]{
   \includegraphics[width=.80\columnwidth]{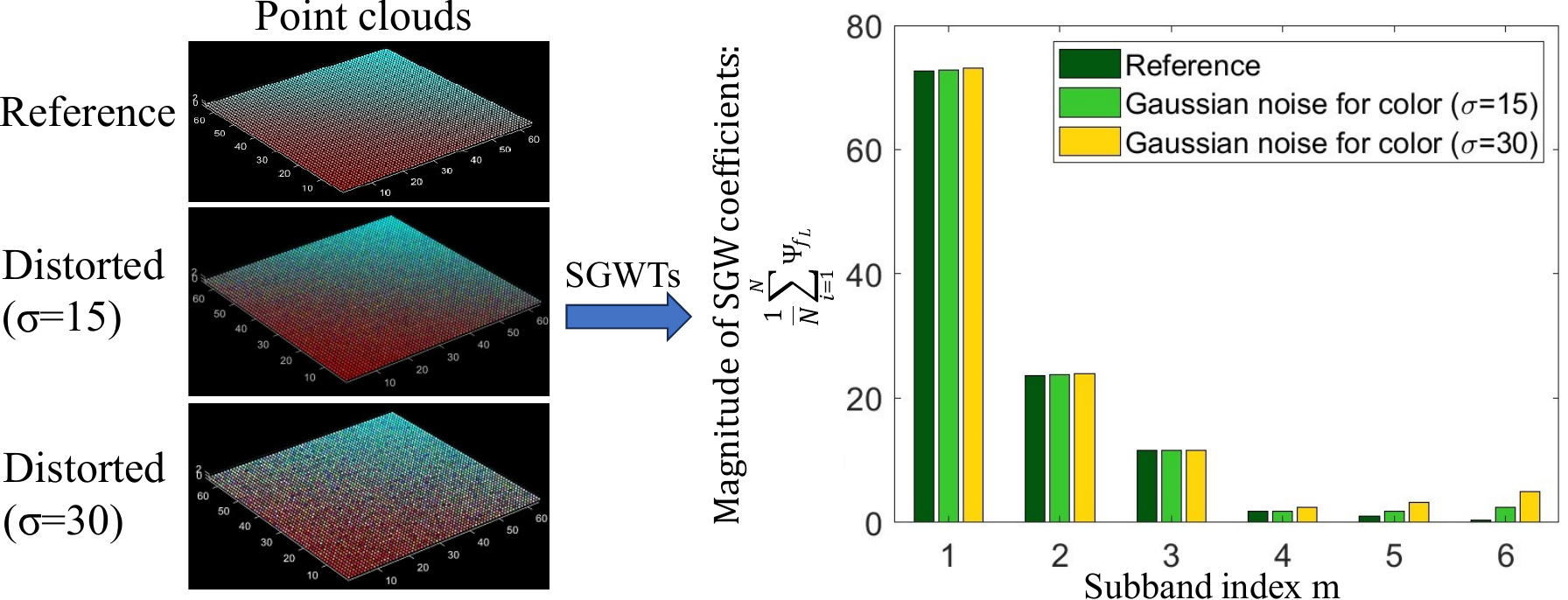}
   \label{fig:graphGaussianColor}
  }
      \subfigure[]{
   \includegraphics[width=.80\columnwidth]{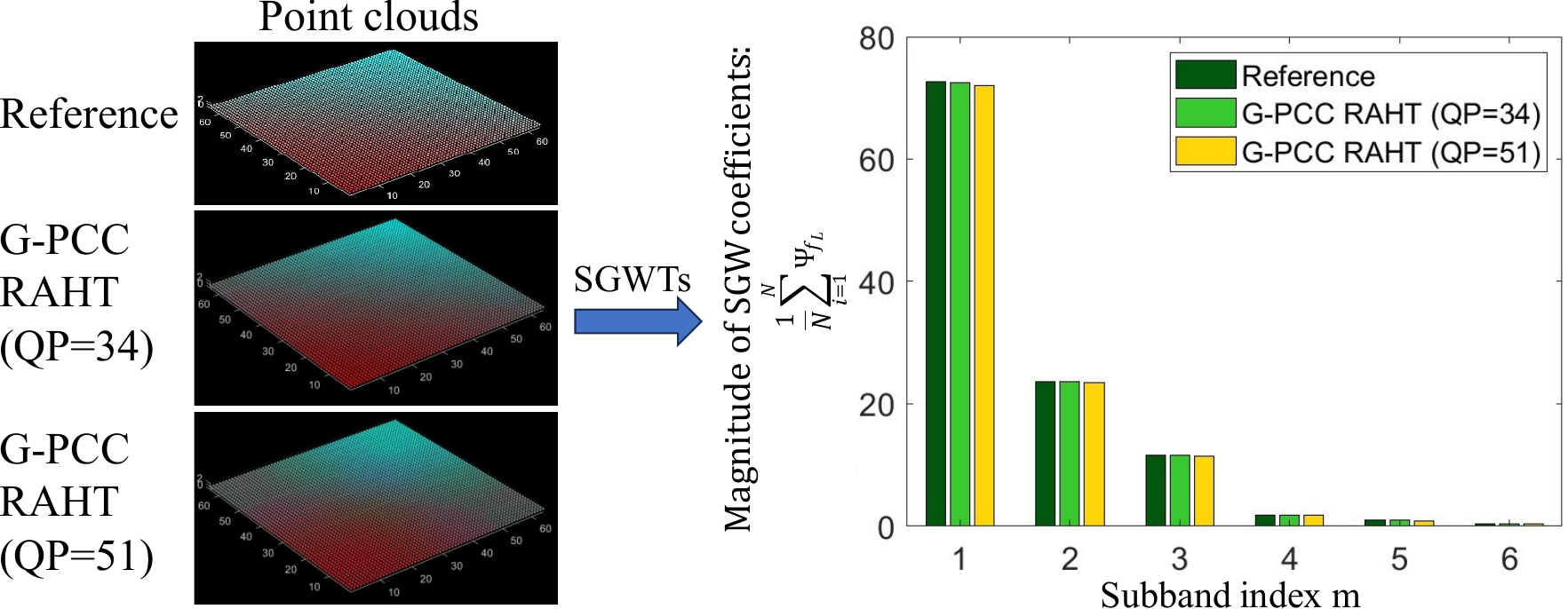}
   \label{fig:graphRAHT}
  }
  \caption{Changes of SGWs through distortions caused by (a) Gaussian noise (geometry), (b) downsampling, (c) Gaussian noise (color), and (d) compression error (G-PCC RAHT \cite{MPEGPCC}).}
  \label{fig:graphChangesSGW}
\end{figure*}
%
%
\section{Preliminaries}
\label{sec:preliminaries}

\subsection{Graph and point cloud notations}
\label{sub-sec:graph-notation}
We begin by introducing mathematical notations and definitions relevant to graphs.
An undirected and weighted graph $\mathcal{G} = (\mathcal{V},\mathcal{E})$ is constructed with a point cloud $\mathrm{P}= \{p_i\}, i=1,...,N$, where $\mathcal{V}$ and $\mathcal{E}$ denote the set of nodes and edges on a graph, respectively.
In this paper, a point cloud $P$ is characterized by the coordinate signals $\bm{f}_g =(\bm{f}_x,\bm{f}_y,\bm{f}_z) \in \mathbb{R}^{N \times 3}$ and color (lightness) signals $\bm{f}_{L} \in \mathbb{R}^N$. As in~\cite{PCQM}, we represent color using lightness in the LAB color space~\cite{LABCOLORSPACE}.

The adjacency matrix $\mathbf{W}= \{ w_{ij} \}$ of the graph corresponding to a point cloud $\mathrm{P}$ defines the edge weights between nodes (points) $i$ and $j$ using the Gaussian kernel function as: 
\begin{equation}
\label{eq:graph-const}
w_{ij}=
\left\{
\begin{array}{ll}
\mathrm{exp} \left( \frac{-||\bm{f}_{g,i} - \bm{f}_{g,j}||_{2}^{2}}{\theta^{2}}\right) & (p_j \in \mathcal{N}(p_i) \textnormal{ or } p_i \in  \mathcal{N}(p_j)) \\
0 & (\mathrm{otherwise})
,
\end{array}
\right.
\end{equation}
where $\bm{f}_{g,i} \in \mathbb{R}^{3}$ shows the coordinate signals at point $p_i$ and $\theta$ denotes the average of all pairwise distances.
We have that $p_i \in \mathcal{N}(p_j)$ if $p_i$ is one of the neighbors of $p_j$.
The definition of neighbors is determined and parameterized by a graph construction algorithm (e.g., \gls{knn}).

The combinatorial graph Laplacian $\mathbf{L}$ is computed by $\mathbf{L} = \mathbf{D} - \mathbf{W}$, where $\mathbf{D}$ is the degree matrix of $\mathbf{W}$ \cite{EmergingGSP}.
$\mathbf{L}$ characterizes the global smoothness of a graph signal $\bm{f} = \{f_i\} \in \mathbb{R}^N$ because the graph Laplacian quadratic form is derived from
%
%
$|\nabla \bm{f} |^2 = \sum_{i \sim j} {w_{ij}(f_i -f_j)^2} = \bm{f} ^\top \mathbf{L} \bm{f},$
%
%
where the sum is over all pairs of connected nodes $i$ and $j$, denoted by $i\sim j$.
The graph Fourier transform $\hat{\bm{f}}$ is defined as $\hat{f}(\lambda_l)=\sum_i f_i\phi_{l,i}$, where $\lambda_l$ and $\phi_{l,i}$ denote the $l$-th eigenvalue and eigenvector of $\mathbf{L}$, respectively.

\subsection{Spectral graph wavelet transform (SGWT)}
\label{sub-sec:sgwt}
Fig.~\ref{fig:sgwt} illustrates the calculation process of \gls{sgwt}~\cite{SGWT,TightFrame} for a point cloud $\mathrm{P}$.
\gls{sgwt}s are constructed using a kernel operator $T_g = g(\mathbf{L})$, which acts on a graph signal $\bm{f}$ by modulating each graph Fourier mode: $\hat{T_g f(l)} = g(\lambda_l)\hat{f}(l)$. 
A scaled operator $T^{m}_{g} = g(S_m\mathbf{L})$ shows the scaling in the spectral domain at scale $S_m$, where $m$ is the index of the band-pass filter, as shown in Fig.~\ref{fig:sgwt}.
At this time, the wavelets are calculated by applying $T^{m}_{g}$ to a single vertex operator: $\psi^i_s=T^{m}_{g}\delta_i$ where $\delta_i$ is the impulse on vertex $i$.
Then, the \gls{sgw} coefficients $\mathbf{\Psi}_f \in \mathbb{R}^{M\times N}$, where $M$ indicates the number of band-pass filters, are calculated by
\begin{equation}
\label{eq:sgwt}
\Psi_f(m,i) = \sum^{N}_{l=1}g(m\lambda_l)\hat{f}(\lambda_l)\phi_{l,i}.
\end{equation}
%
%
%
%
%
\section{Proposed method}
\label{sec:proposed-method}
\subsection{Overview of the proposed method}
\label{sub-sec:proposed-method-overview} 
Reference and distorted point clouds are represented as $\mathrm{P_R} = \{p^{R}_i\},$ $i=1,...,|\mathrm{P_R}|$ and $\mathrm{P_D} = \{p^{D}_j \},$ $ j=1,...,|\mathrm{P_D}|$, respectively, where $|\mathrm{P_R}|$ denotes the number of points of $\mathrm{P_R}$.
Point clouds $\mathrm{P_R}$ and  $\mathrm{P_D}$ have three coordinate signals and one color signal $(\bm{f}^R_x,\bm{f}^R_y,\bm{f}^R_z,\bm{f}^R_L)$ and $(\bm{f}^D_x,\bm{f}^D_y,\bm{f}^D_z,\bm{f}^D_L)$, respectively.
Fig.~\ref{fig:flow} shows the overall calculation flow of the proposed method called \gls{sgwpcqa}, which includes the following five steps.
%
%
\begin{enumerate}
\setlength{\itemsep}{0cm}
\setlength{\leftskip}{-0.4cm}
\item 
A graph $\mathcal{G}_R$ with adjacency matrix $\mathbf{W_R}$ is constructed from $\mathrm{P_R}$ by \gls{knn}.
\item 
Point-to-point correspondence between $\mathrm{P_R}$ and $\mathrm{P_D}$ is calculated. The nearest point in $\mathrm{P_D}$ from $p^R_i$ is selected as $\Tilde{p}^D_i$  by the nearest neighbor search using Euclidean distance.
The associated distorted point cloud $\mathrm{\Tilde{P}_D} = \{\Tilde{p}^D_i\}$ has the signals $(\Tilde{\bm{f}}^D_x,\Tilde{\bm{f}}^D_y,\Tilde{\bm{f}}^D_z,\Tilde{\bm{f}}^D_L)$.
\item 
The \gls{sgw} coefficients  with graph $\mathcal{G}_R$ are calculated for each graph signal $\bm{f}^R_x$, $\bm{f}^R_y$, $\bm{f}^R_z$, $\bm{f}^R_L $, $ \Tilde{\bm{f}}^D_x $, $\Tilde{\bm{f}}^D_y$, $\Tilde{\bm{f}}^D_z$ and $\Tilde{\bm{f}}^D_L$. The resulting \gls{sgw} coefficients are $\bm{\Psi}^R_{f_x}$, $\bm{\Psi}^R_{f_y}$ ,$\bm{\Psi}^R_{f_z}$, $\bm{\Psi}^R_{f_L}$, $ \Tilde{\bm{\Psi}}^D_{f_x}$, $\Tilde{\bm{\Psi}}^D_{f_y}, \Tilde{\bm{\Psi}}^D_{f_z}$, and $\Tilde{\bm{\Psi}}^D_{f_L}$.
 
\item
The geometry \gls{pcqa} score $S^{G}_m$ and color \gls{pcqa} score $S^{G}_m$ are computed by comparing the \gls{sgw} coefficients at the $m$-th subband.
\item A trained \gls{svr} model predicts a final assessment score from the scores $S^{G}_m$ and $S^{C}_m$. Note that the parameters of the \gls{svr} model are optimized using a training set, including the scores calculated in step 4 and the \gls{mos} of the corresponding distorted point clouds.
\end{enumerate}
%
%
We explain steps 1, 2, and 3 in  \autoref{sub-sec:proposed-sgwt}.
Steps 4 and 5 are described in  \autoref{sub-sec:sgwt-pcqa} and \autoref{sub-sec:svr}, respectively.
Furthermore, \gls{sgwpcqa}+, which is an extension of \gls{sgwpcqa}, is described in \autoref{sub-sec:sgwpcqa+}.
\subsection{Calculation of spectral graph wavelets}
\label{sub-sec:proposed-sgwt} 

This section explains how to calculate SGWs for PCQA. In graph signal processing, the definition of graph frequency depends on the structure of the graph \cite{EmergingGSP}.
Hence, if graphs are independently constructed from reference and distorted point clouds, the definitions of their respective graph frequencies will be different.
Therefore, to perform \gls{pcqa}, if we wish to compare the graph frequencies obtained from $P_R$ and $P_D$, we first must select a common graph for the two point clouds.  

In the proposed method, a graph $\mathcal{G}_R$ with the adjacency matrix $\mathbf{W_R}$ is calculated from the reference point cloud $\mathrm{P_R}$ by \gls{knn}, using \eqref{eq:graph-const}. 
Then, the nearest point $\Tilde{p}^D_i \in \mathrm{P_D}$ from a query point $p^R_{i}$ is extracted by the nearest neighbor search based on the 3D Euclidean distance to project $\mathrm{P_D}$ into $\mathrm{P_R}$.
In this way, we obtain a new distorted point cloud $\mathrm{\Tilde{P}_D} = \{\Tilde{p}^D_i\}$ that has the same node indices but different graph signals. 
Finally, the \gls{sgw} coefficients are calculated from coordinate and color signals of $\mathrm{P_R}$ and $\mathrm{\tilde{P}_D}$ based on \eqref{eq:sgwt}.
Since \gls{sgwt}s are computed for each graph signal ($\bm{f}^R_x,\bm{f}^R_y,\bm{f}^R_z,\bm{f}^R_L, \Tilde{\bm{f}}^D_x,\Tilde{\bm{f}}^D_y,\Tilde{\bm{f}}^D_z$ and $\Tilde{\bm{f}}^D_L$), SGWT coefficients ($\bm{\Psi}^R_{f_x}, \bm{\Psi}^R_{f_y}, \bm{\Psi}^R_{f_z}, \bm{\Psi}^R_{f_L}, \Tilde{\bm{\Psi}}^D_{f_x},\Tilde{\bm{\Psi}}^D_{f_y},\Tilde{\bm{\Psi}}^D_{f_z}$, and $\Tilde{\bm{\Psi}}^D_{f_L}$) are output.

\subsection{SGW-based FR-PCQA metrics}
\label{sub-sec:sgwt-pcqa}

Fig.~\ref{fig:graphChangesSGW} shows the results of our preliminary experiment, which confirmed that SGWs change under several types of distortions.
As shown in Fig.~\ref{fig:graphChangesSGW}, since \gls{sgw}s at various subbands are changed when distortions are introduced, the quality can be assessed by observing the changes of \gls{sgw}s.

In the proposed \gls{sgwpcqa} method, to assess geometric distortion, the differences in SGWs of $x,y$, and $z$ coordinates for each $m$-th subband are calculated as follows:
\begin{equation}
\label{eq:x-pcqa}
G^{x}_m = \frac{1}{|P_R|} \sum_{p^R_i \in \mathrm{P_R}} (\Psi^R_{f_x}(m,i) - \Tilde{\Psi}^D_{f_x}(m,i))^2,
\end{equation}
\begin{equation}
\label{eq:y-pcqa}
G^{y}_m = \frac{1}{|P_R|} \sum_{p^R_i \in \mathrm{P_R}} (\Psi^R_{f_y}(m,i) - \Tilde{\Psi}^D_{f_y}(m,i))^2,
\end{equation}
\begin{equation}
\label{eq:z-pcqa}
G^{z}_m = \frac{1}{|P_R|} \sum_{p^R_i \in \mathrm{P_R}} (\Psi^R_{f_z}(m,i) - \Tilde{\Psi}^D_{f_z}(m,i))^2,
\end{equation}
\begin{equation}
\label{eq:geom-pcqa}
G_m = \frac{G^{x}_m + G^{y}_m + G^{z}_m}{3}.
\end{equation}
Geometric assessment scores $S^{G}_m$ are calculated as
\begin{equation}
\label{eq:score-geom-pcqa}
S^{G}_m = \frac{1}{1+G_m}.
\end{equation}
Likewise, color assessment scores $S^{C}_m$ are given as 
\begin{equation}
\label{eq:L-pcqa}
C_m = \frac{1}{|P_R|} \sum_{p^R_i \in \mathrm{P_R}} (\Psi^R_{f_L}(m,i) - \Tilde{\Psi}^D_{f_L}(m,i))^2,
\end{equation}
\begin{equation}
\label{eq:score-geom-pcqa}
S^{C}_m = \frac{1}{1+C_m}.
\end{equation}
\subsection{Score integration by support vector regression (SVR)}
\label{sub-sec:svr}
We use \gls{svr} to predict a final \gls{pcqa} score.
Multiple scores, $S^{G}_m$ and $S^{C}_m$, are integrated by \gls{svr}.
During the training phase, an \gls{svr} model is trained using the \gls{mos} of the corresponding distorted point clouds and the multiple \gls{pcqa} scores calculated from training data.
The Gaussian radial basis function (RBF) kernel and the sequential minimal optimization \cite{SMO} are chosen as a kernel and solver of \gls{svr}, respectively.

\subsection{Extension of \gls{sgwpcqa} (\gls{sgwpcqa}+)}
\label{sub-sec:sgwpcqa+}
In our preliminary experiments, the performance of \gls{sgwpcqa} can be further improved by adding some of the \gls{fr-pcqa} scores utilized in our previous work, \gls{frsvr} \cite{ICIPChallenge}.
Thus, we selected two types of scores utilized in \gls{frsvr}, (1) point-to-point score ($S_\mathrm{p2p}$) and (2) \gls{gtv}-based score ($S_\mathrm{gtv}$), whose addition significantly improved the correlation with subjective scores.
The point-to-point score $S_\mathrm{p2p}$ is based on point-to-point error \cite{MPEGEval}.
The \gls{gtv}-based score
 $S_\mathrm{gtv}$ is calculated by comparing \gls{gtv}s, which quantify the smoothness of graph signals (Please see \cite{ICIPChallenge} to access the detailed formulas due to space limitation).

Although \gls{sgw}s reflect graph signal variation at various scales, they cannot consider too local and global information. Since the point-to-point and \gls{gtv}-based scores focus on very local (point-wise) and global information, respectively, adding these scores leads to improved correlation with subjective scores.
In addition, although graph construction is required to calculate these two scores, a graph utilized for calculating \gls{sgw}s can be reused. Thus, the additional processing time is not very large.
The extension that combines these two scores ($S_\mathrm{p2p}$ and $S_\mathrm{gtv}$) with the SGW-based scores ($S^{G}_m$ and $S^{C}_m$) using SVR is named as \gls{sgwpcqa}+.
%

%
\section{EXPERIMENTS}
\label{sec:experiments}
\begin{figure}[t]
\centering 
   \subfigure[]{
   \includegraphics[width=.40\columnwidth]{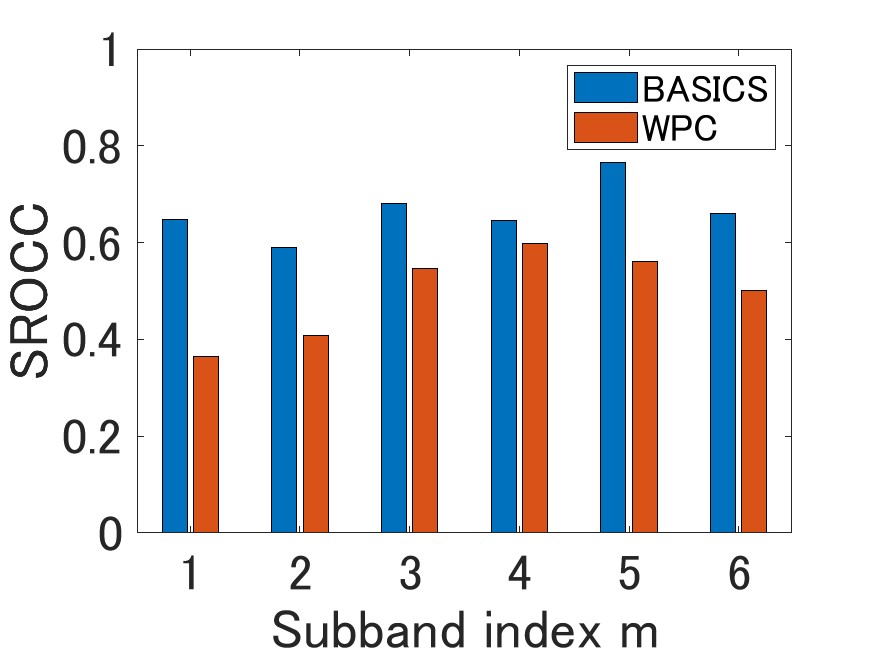}
   \label{fig:geomGraph}
  }
  \subfigure[]{
   \includegraphics[width=.40\columnwidth]{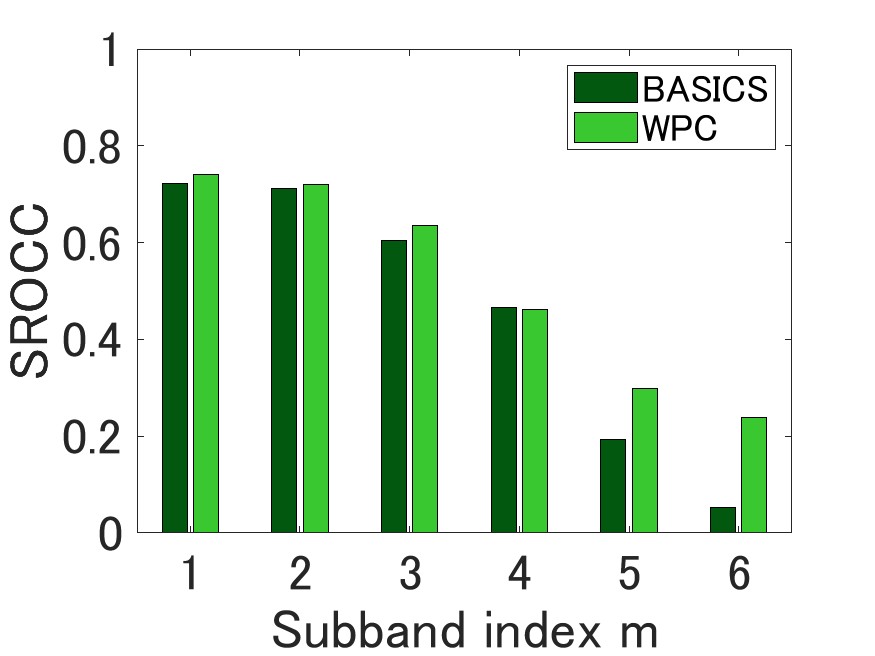}
   \label{fig:colorGraph}
  }
  \caption{SROCCs of each \gls{sgwpcqa} score, (a) geometry-based score ($S^G_m$), and (b) color-based score ($S^C_m$), with the BASICS dataset (test set).}
  \label{fig:eachScoreSROCC}
\end{figure}
\begin{table*}[t]
\centering
\small
\caption{The correlation coefficients of the proposed \gls{sgwpcqa} method. Some of the scores were selected to train an \gls{svr} model. The \textbf{\color{blue}blue} characters mean the best result.}
\begin{tabular}{ccccccccc}
\hline
\multirow{2}{*}{\#} & \multirow{2}{*}{Geometry score set} & \multirow{2}{*}{Color score set} & \multicolumn{2}{c}{BASICS \cite{BASICS}} & \multicolumn{2}{c}{WPC \cite{WPC}} & \multicolumn{2}{c}{Average} \\ \cline{4-9} 
                    &                                     &                                  & PLCC         & SROCC       & PLCC       & SROCC      & PLCC         & SROCC        \\ \hline
\#1 & $S^G_1$ & $S^C_1$  & 0.844 & 0.770 & 0.787 & 0.780 & 0.816 & 0.775 \\
\#2 & $S^G_1$, $S^G_2$ & $S^C_2$,$S^C_2$ & 0.880 & 0.815 & 0.835 & 0.815 & 0.858 & 0.815 \\
\#3 & $S^G_1$, $S^G_2$, $S^G_3$ & $S^C_1$, $S^C_2$, $S^C_3$ & 0.874 & 0.837 & 0.854 & 0.842 & 0.864 & 0.840 \\
\#4 & $S^G_1$, $S^G_2$, $S^G_3$, $S^G_4$ & $S^C_1$, $S^C_2$, $S^C_3$, $S^C_4$  & 0.870 & 0.837 & \textbf{\color{blue}0.863} & \textbf{\color{blue}0.854} & 0.867 & 0.846 \\
\#5 & $S^G_1$, $S^G_2$, $S^G_3$, $S^G_4$, $S^G_5$ & $S^C_1$, $S^C_2$, $S^C_3$, $S^C_4$, $S^C_5$ & 0.875 & 0.835 & 0.847 & 0.834 & 0.861 & 0.835     \\
\#6 & $S^G_1$, $S^G_2$, $S^G_3$, $S^G_4$, $S^G_5$, $S^G_6$ & $S^C_1$, $S^C_2$, $S^C_3$, $S^C_4$, $S^C_5$, $S^C_6$ & 0.889 & 0.843 & 0.826 & 0.814 & 0.858 & 0.829 \\ \hline
\#7 & $S^G_1$, $S^G_2$, $S^G_3$, $S^G_4$, $S^G_5$, $S^G_6$ & $S^C_1$, $S^C_2$, $S^C_3$ & \textbf{\color{blue}0.892} & \textbf{\color{blue}0.861} & 0.845 & 0.840 & \textbf{\color{blue}0.869} & \textbf{\color{blue}0.851}  \\
\#8 & $S^G_1$, $S^G_2$, $S^G_3$, $S^G_4$, $S^G_5$, $S^G_6$ & $S^C_1$, $S^C_2$, $S^C_3$, $S^C_4$ & 0.888 & 0.859 & 0.842 & 0.837 & 0.865 & 0.848  \\
\#9 & $S^G_1$, $S^G_2$, $S^G_3$, $S^G_4$, $S^G_5$, $S^G_6$ & $S^C_1$, $S^C_2$, $S^C_3$, $S^C_4$,  $S^C_5$  & 0.879        & 0.835       & 0.828      & 0.823      & 0.854        & 0.829       \\ \hline
\end{tabular}
\label{tab:prop-comparison}
\end{table*}
\begin{table*}[]
\centering
\small
\caption{The correlation coefficients and processing time [s] of the proposed and conventional methods. The ``Average'' indicates the average of the two datasets. The \textbf{\color{blue}blue} and \textbf{\color{red}red} characters mean the first and second best results, respectively.}
\begin{tabular}{lccccccccc}
\hline
\multirow{2}{*}{Metric} & \multicolumn{3}{c}{BASICS~\cite{BASICS}}   & \multicolumn{3}{c}{WPC~\cite{WPC}}      & \multicolumn{3}{c}{Average}  \\ \cline{2-10} 
                        & PLCC  & SROCC & Time {[}s{]} & PLCC  & SROCC & Time {[}s{]} & PLCC  & SROCC & Time {[}s{]} \\ \hline
Point-to-point (MSE) \cite{MPEGEval}    & 0.041 & 0.735 & \textbf{\color{blue}{9.701}}        & 0.401 & 0.566 & \textbf{\color{blue}{6.832}}        & 0.221 & 0.651 & \textbf{\color{blue}{8.267}}        \\
Point-to-plane (MSE)  \cite{MPEGEval}   & 0.005 & 0.799 & 49.122       & 0.368 & 0.481 & 32.847       & 0.187 & 0.640 & 40.985       \\
Angular similarity \cite{AngularSimilarity}     & 0.328 & 0.306 & 39.155       & 0.296 & 0.319 & 27.336       & 0.312 & 0.313 & 33.246       \\
Y-MSE \cite{MPEGEval}                   & 0.512 & 0.550 & \textbf{\color{red}{9.896}}        & 0.469 & 0.591 & \textbf{\color{red}{6.928}}        & 0.491 & 0.571 & \textbf{\color{red}{8.412}}        \\
PointSSIM  \cite{PointSSIM}       & 0.605 & 0.620 & 23.682       & 0.469 & 0.471 & 15.281       & 0.537 & 0.546 & 19.482       \\
PCQM \cite{PCQM}                  & 0.786 & 0.739 & 256.463      & 0.511 & 0.550 & 214.197      & 0.649 & 0.645 & 235.330      \\
GraphSIM \cite{GraphSIM}                & 0.801 & 0.773 & 613.652      & 0.688 & 0.691 & 382.798      & 0.745 & 0.732 & 498.225      \\
MSGraphSIM \cite{MSGraphSIM}            & 0.807 & 0.773 & 645.589      & 0.712 & 0.724 & 348.183      & 0.760 & 0.749 & 496.886      \\
FRSVR \cite{ICIPChallenge}                   &   \textbf{\color{red}{0.914}}    &  \textbf{\color{red}{0.878}}     &    32.722          &    0.818   &   0.803    &    20.608          &   0.866   &   0.841    &      26.665        \\ \hline
SGW-PCQA (Proposed, \#7 in Table \ref{tab:prop-comparison})               & 0.892     & 0.861     &   27.387         & \textbf{\color{red}{0.845}}    & \textbf{\color{red}{0.840}}     & 15.526            & \textbf{\color{red}{0.869}}     & \textbf{\color{red}{0.851}}     &     21.457   \\     
SGW-PCQA+ (Proposed)                & \textbf{\color{blue}{0.924}}     & \textbf{\color{blue}{0.888}}     &   30.082          & \textbf{\color{blue}{0.873}}    & \textbf{\color{blue}{0.869}}     & 17.266            & \textbf{\color{blue}{0.899}}     & \textbf{\color{blue}{0.879}}     &      23.674       \\ \hline
\end{tabular}
\label{tab:conv-comparison}
\end{table*}
\subsection{Experimental conditions}
\label{sub-sec:experiments-conditions}
\noindent \textbf{Evaluation Criteria:} To quantify the correlation to the subjective assessment score, we employed Pearson’s linear correlation coefficient (PLCC) and Spearman’s rank-order correlation coefficient (SROCC). 

\noindent \textbf{Dataset:} We used the two datasets, the \gls{basics}~\cite{BASICS}, consisting of 1498 point clouds distorted by compression errors, and the \gls{wpc}~\cite{WPC} datasets, consisting of 740 distorted point clouds contaminated by compression, Gaussian noise, or down-sampling.
The two datasets include the \gls{mos} of the corresponding distorted point clouds to be utilized to measure PLCC and SROCC.
Since the \gls{basics} dataset was explicitly divided into training and test sets~\cite{BASICS}, we conducted training and evaluation with the training and test sets, respectively. 
As for the \gls{wpc} dataset, distorted point clouds were segmented into five parts (20\% data $\times$ 5 parts). 
Each segment served as an independent test set, while the remaining data were utilized as training data. 
After calculating PLCC and SROCC for each test set, the average was shown as the evaluation result.

\noindent \textbf{Machine specifications:} The processing time was measured with a computer equipped with an AMD Ryzen Threadripper 2970WX 24-Core processor, NVIDIA GTX 1080 Ti, and 128GB Random Access Memory. 

\noindent \textbf{Implementation details:} 
The graph construction parameter $K$ and the number of wavelet decomposition levels $M$ were set to 8 and 6, respectively.
Since many of the conventional methods compared in this paper do not employ GPU computing for acceleration, GPU computing is not utilized for the implementation of the proposed method for a fair comparison.
\subsection{Experiment 1: Performance of a single score}
\label{sub-sec:exp1}

We evaluated the performance of a single \gls{sgwpcqa} score, $S^G_m$ and $S^C_m$ (see Fig.\ref{fig:eachScoreSROCC}).
Since the SROCCs were calculated from only one score in this experiment, \gls{svr} was not utilized for integrating multiple scores.
As shown in Fig.~\ref{fig:colorGraph}, the SROCC of color-based score $S^C_m$ was lower for the high-frequency subbands.
For the high frequencies of color information, there were many cases where the differences between reference and distorted point clouds were small.
We observed that the color signals of a reference point cloud had very few high-frequency components, even for general point clouds other than a simple point cloud shown in Fig.\ref{fig:graphRAHT}.
Since small high-frequency components are quantized to zero, the compression process does not add high-frequency components.
Comparing high-frequency components is not effective in compression noise of color signals, which leads to low SROCCs.
Thus, the SROCC of the \gls{basics} dataset consisting only of compression errors was particularly low compared to that of the \gls{wpc} dataset.

\subsection{Experiment 2: Performance of the combination of multiple scores}
\label{sub-sec:exp2}

We investigated the performance achieved by combining multiple scores using \gls{svr}.
\autoref{tab:prop-comparison} shows the PLCC and SROCC of the two datasets.
Condition \#7 in \autoref{tab:prop-comparison}, where all geometry scores and three color scores ($S^C_1$, $S^C_2$, and $S^C_3$) were selected, recorded the best average PLCC and SROCC. 
Since the SROCCs of color-based scores at high-frequency bands were relatively low as shown in Fig.~\ref{fig:eachScoreSROCC}, using the color-based scores $S^C_m$ at high-frequency bands harmed the correlation with subjective assessment scores.

\subsection{Experiment 3: comparison with SOTA methods}
\label{sub-sec:exp3}

We compared the proposed method with conventional \gls{fr-pcqa} methods~ \cite{MPEGEval,C2PError,PointSSIM,PCQM,GraphSIM,MSGraphSIM,ICIPChallenge,AngularSimilarity}.
\autoref{tab:conv-comparison} presents the results of the comparison experiment with PLCC, SROCC, and average processing time across all the point clouds.
\gls{sgwpcqa} and \gls{sgwpcqa}+ in \autoref{tab:conv-comparison} adopted condition \#7 in \autoref{tab:prop-comparison} because it achieved the best results in \autoref{sub-sec:exp2}.
The results demonstrate that the proposed \gls{sgwpcqa}+ achieved the highest PLCC and SROCC. 
In particular, the PLCC and SROCC of the WPC dataset were improved against our previous study, \gls{frsvr} \cite{ICIPChallenge}.
The \gls{pcqa} using \gls{sgw}s were effective with various noise, such as Gaussian noise and down-sampling.
For compression noise, the five metrics adopted in \gls{frsvr} \cite{ICIPChallenge} provided sufficiently high accuracy; however, those metrics can be insufficient for diverse types of noise.
In addition, since \gls{sgwt}s are calculated relatively fast by polynomial approximation~\cite{SGWT}, \gls{sgwpcqa} performed faster computation compared to conventional multiscale metric (e.g., MSGraphSIM~\cite{MSGraphSIM}).

\section{CONCLUSION}
\label{sec:conclusion}
This paper introduced an accurate \gls{fr-pcqa} method using \gls{sgw}s.
The proposed metric demonstrates a great correlation by referring to the \gls{sgw}s calculated from the coordinate and color signals. 
In addition, we made the method more accurate by integrating our previous method called \gls{frsvr} with the \gls{sgwpcqa} metrics by using \gls{svr}.
In the future, we will consider a more suitable graph construction method for \gls{pcqa} instead of \gls{knn}, which is adopted in the proposed method. 
Since the previous denoising study \cite{SCNNK} has shown that the values of \gls{sgw}s highly depend on the graph structure, there may be a more suitable graph construction for \gls{pcqa}.

\section*{\small ACKNOWLEDGEMENTS}
This work was supported by the Ministry of Internal Affairs and Communications (MIC) of Japan (Grant no. JPJ000595).

\newpage
\footnotesize
\bibliographystyle{IEEEbib}
\bibliography{ICIP2024}

\end{document}